\documentclass[conference]{IEEEtran}
\IEEEoverridecommandlockouts
\usepackage{cite}
\usepackage{amsmath,amssymb,amsfonts}
\usepackage{graphicx}
\usepackage{textcomp}
\usepackage{xcolor}
\usepackage{algorithm}
\usepackage{algpseudocode}
\usepackage{amsmath}
\usepackage{url}
\def\BibTeX{{\rm B\kern-.05em{\sc i\kern-.025em b}\kern-.08em
    T\kern-.1667em\lower.7ex\hbox{E}\kern-.125emX}}
\begin{document}


\title{Toward Knowledge-Driven Speech-Based Models of Depression: Leveraging Spectrotemporal Variations in Speech Vowels
\thanks{This work is supported by the National Science Foundation (CAREER: Enabling Trustworthy Speech Technologies for Mental Health Care: From Speech Anonymization to Fair Human-centered Machine Intelligence, \#2046118). The code developed as part of this work is publicly available at
\protect\url{https://github.com/HUBBS-Lab-TAMU/2dCNN-LSTM-depression-identification}
}}

\author{\IEEEauthorblockN{Kexin Feng}
\IEEEauthorblockA{\textit{Computer Science and Engineering} \\
\textit{Texas A\&M University}\\
kexin@tamu.edu}
\and
\IEEEauthorblockN{Theodora Chaspari}
\IEEEauthorblockA{\textit{Computer Science and Engineering} \\
\textit{Texas A\&M University}\\
chaspari@tamu.edu}
}

\maketitle

\begin{abstract}
Psychomotor retardation associated with depression has been linked with tangible differences in vowel production. This paper investigates a knowledge-driven machine learning (ML) method that integrates spectrotemporal information of speech at the vowel-level to identify the depression. Low-level speech descriptors are learned by a convolutional neural network (CNN) that is trained for vowel classification. The temporal evolution of those low-level descriptors is modeled at the high-level within and across utterances via a long short-term memory (LSTM) model that takes the final depression decision. A modified version of the Local Interpretable Model-agnostic Explanations (LIME) is further used to identify the impact of the low-level spectrotemporal vowel variation on the decisions and observe the high-level temporal change of the depression likelihood. The proposed method outperforms baselines that model the spectrotemporal information in speech without integrating the vowel-based information, as well as ML models trained with conventional prosodic and spectrotemporal features. The conducted explainability analysis indicates that spectrotemporal information corresponding to non-vowel segments less important than the vowel-based information. Explainability of the high-level information capturing the segment-by-segment decisions is further inspected for participants with and without depression. The findings from this work can provide the foundation toward knowledge-driven interpretable decision-support systems that can assist clinicians to better understand fine-grain temporal changes in speech data, ultimately augmenting mental health diagnosis and care.
\end{abstract}

\begin{IEEEkeywords}
Mental health, depression, speech vowels, convolutional neural network (CNN), long short-term memory (LSTM), explainable machine learning
\end{IEEEkeywords}

\section{Introduction}
Depression is the most common mental health (MH) condition influencing approximately 280 million people worldwide~\cite{who_depression}. Patients with depression experience among others feelings of sadness, irritability, and emptiness, loss of pleasure or interest in activities, and feelings of guilt or low self-worth. Despite the high prevalence depression and its impact on patients’ functioning and quality of life, only approximately half of respondents with depressive or anxiety disorders receive treatment~\cite{katon2013depression} due to lack of insurance coverage, unequal access to evidence-based practices, stigma, MH workforce shortages, and geographical maldistribution of providers. These challenges are particularly prevalent amongst racial-ethnic minority groups and underserved communities that also frequently suffer from diagnostic assessment bias and diagnostic errors~\cite{mongelli2020challenges}.

Qualitative and quantitative evidence suggests that speech production mechanisms are influenced by depression, and that patients with depression depict slowed speech rate, monotonous pitch, and reduced loudness, which is reflected in prosodic measures, such as fundamental frequency (F0), intensity, and speaking rate~\cite{mundt2007voice}. In addition, speech technologies in tandem with machine learning (ML) can be ubiquitous, therefore rendering access to in-situ ecologically valid data and allowing just-in-time support during states of opportunity and states of vulnerability~\cite{spruijt2014dynamic}. Yet, converting ML-derived decisions into effective action remains a challenge~\cite{james2022preparing}. Clinicians may find it difficult to trust complex ML algorithms over their own intuition, unless they are provided with explanations that can facilitate alert interpretation and promote the transparency of the ``black-box" system~\cite{ginestra2019clinician}.

Depression can influence the motor control and consequently, speech production. Speech is produced as a sound generated by the glottis and modulated by the vocal tract which acts as a resonant filter, causing formant frequencies and spectrotemporal variations~\cite{atal1982new}. Physiological and cognitive impairments associated with MH affect the phonological loop~\cite{murphy1999emotional} and produce noticeable energy variations in speech vowels. An early study by Shimizu {\it et al.} suggests that people with depression depict decreased laryngeal vagal function, which controls the chaotic motion of the vocal folds, and thus, is manifested as an increased chaotic pattern in vowel sounds \cite{shimizu2005chaos}. Scherer {\it et al.} found that individuals with depression depict reduced vowel space, defined as the frequency range between the first and second formant (i.e., F1 and F2) of the vowels, compared to their healthy counterparts \cite{scherer2015self}. Vlasenko {\it et al.} analyzed the effect of depression on formant dynamics for female and male speakers separately \cite{vlasenko2017implementing}. Results demonstrate the potential of using gender-dependent vowel-based features for depression detection, which outperformed a range of turn-level acoustic features that were extracted without considering vowel-based information. Finally, Stasak {\it et al.} examined linguistic stress differences between non-depressed and clinically depressed individuals and found statistically significant differences in vowel articulatory parameters with shorter vowel durations and less variance for 'low', 'back', and 'rounded' vowel positions for individuals with depression \cite{stasak2019investigation}. Evidence from these studies suggests that the effects of psychomotor retardation associated with depression are linked to tangible differences in vowel production, whose measures can be particularly informative when detecting depression from speech.

Leveraging this evidence, we proposed a vowel-dependent deep learning approach for classifying depression from speech. We extract low-level vowel-dependent feature embeddings via a convolutional neural network (CNN), which is trained for the task of vowel classification, thus, capturing energy variations of the speech spectrogram within the corresponding vowel region. The learned vowel-based embeddings comprise the input to a long short-term memory (LSTM) model that models high-level temporal dependencies across speech segments and outputs the depression outcome. We compare the proposed approach against prior work on deep learning models that does not leverage vowel-dependent information and find that the proposed method depicts better performance compared to the considered baselines. We further conduct an empirical analysis using a modified version of the Local Interpretable Model-agnostic Explanations (LIME) to identify the parts of the speech spectrogram that contribute most to the decision about the depression condition. Results from this work could potentially contribute to reliable and interpretable speech-based ML models of depression that consider low-level fine-grain spectrotemporal information along with high-level transitions within and across an utterance, and could be eventually used to assist MH clinicians in decision-making.

\section{Previous work} \label{Sec: previous}
Early work in depression estimation from speech has explored acoustic biomarkers of depression that quantify characteristics of the vocal source, tract, formants, and prosodics. Psychomotor retardation associated with depression can slow the movement of articulatory muscles, therefore causing decreases in speech rate and phoneme rate, as well as decreases in pitch and energy variability \cite{mundt2007voice,quatieri2012vocal}. Other work has quantified changes in coordination of the vocal tract motion by measuring correlation coefficients over different temporal scales across formant frequencies and the first-order derivatives of the Mel-frequency cepstral coefficients (MFCC) \cite{williamson2013vocal}. Formant information has been considered by measuring the distance between F1 and F2 (the first and second formant) coordinates in certain vowels (i.e., /i/, /u/, /a/), and showing that this distance is reduced for patients with depression compared to healthy individuals \cite{scherer2015self}. Finally, other work has modeled the acoustic variability of MFCCs at the utterance-level indicating the patients with depression depict reduced variability in regards to this spectral measure~\cite{cummins2015analysis}. The above measures along with commonly used acoustic features of prosody, energy, intonation, and spectral information have been employed as the input to various ML models for the task of depression recognition~\cite{valstar2013avec, cohn2018multimodal}. 

Other work has leveraged recent advances in deep learning to reliably estimate depression from speech. Ma {\it et al.} proposed a deep model, called ``DepAudioNet" that encoded the Mel-scale filter bank of speech via a 1-dimensional CNN followed by a LSTM \cite{ma2016depaudionet}. The convolutional transformations implemented in the 1D CNN encode low-level short-term spectrotemporal speech variations, the max-pooling layers of the 1D CNN capture mid-level variations, while the LSTM extracts long-term information within an utterance. Muzammel {\it et al.} incorporated speech spectrograms at the phoneme-level as the input to 2-dimensional CNNs \cite{muzammel2020audvowelconsnet}. Vowels and consonants were separated from speech signals and served as the input to a 2-D CNN that performed depression classification. Results demonstrate that the fusion of consonant-based and vowel-based embeddings depict the best performance. Saidi {\it et al.} used the speech spectrogram as the input of a 2-D CNN that was trained for depression classification \cite{saidi2020hybrid}. The learned embeddings of the CNN were flattened and comprised the input to a support vector machine (SVM). The CNN-SVM combination depicted an absolute of 10\% increase in depression classification accuracy compared to the CNN alone. Srimadhur \& Lalitha compared a 2-D CNN whose input was the speech spectrogram with an end-to-end 1D CNN whose input was the raw speech signal, with the latter yielding improved results \cite{srimadhur2020end}. Zhao {\it et al.} proposed the hierarchical attention transfer network (HATN), a two-level hierarchical network with an attention mechanism that models speech variations at the frame and sentence level \cite{zhao2020hierarchical}. This model learns attention weights for a speech recognition task (source task), implemented via a deep teacher network, and transfers those weights to the depression severity estimation task (target task), implemented via a shallow student network.

The contributions of this work compared to the previous research are: (1) In contrast to the majority of prior work in deep learning, this paper proposes a CNN-based architecture that learns vowel-based embeddings, therefore explicitly incorporating low-level spectrotemporal information that is valuable for identifying depression \cite{shimizu2005chaos, scherer2015self, scherer2015reduced, vlasenko2017implementing, stasak2019investigation}. While phoneme-based information has been modeled before \cite{muzammel2020audvowelconsnet}, it was only incorporated at the extraction of spectrogram features, without further deriving high-level representations; (2) High-level information is extracted from the vowel-based embeddings via the LSTM that models the evolution of this information within and across utterances, a technique that is fairly unexplored in depression detection and has not been combined with vowel-based information~\cite{ma2016depaudionet}; and (3) The utility of learned embeddings is qualitatively explored in terms of its interpretability with respect to the depression outcome.

\section{Proposed Methodology}
The proposed method will learn low-level descriptors of spectrotemporal variations around speech vowels, namely, ``vowel-based embeddings". Following that, the vowel-based embeddings will serve as the input to a LSTM which will model high-level temporal dependencies within and across utterances. These will be achieved via the following steps: (1) vowel segmentation module, that splits an utterance into segments and assigns a vowel class to each segment (Section~\ref{sssec:Vowelseg}); (2) vowel classification module, that uses the aforementioned speech segments to conduct a vowel classification task, while also learning vowel-based embeddings from speech (Section~\ref{sssec:Vowelclass}); and (3) depression classification module, that takes as an input the sequence of learned vowel-based embeddings from the previous task and learns their temporal evolution in association to the depression outcome (Section~\ref{sssec:Depressionclass}). We will consider five English vowels (e.g., /a/, /e/, /i/, /o/, /u/) produced through different vocal tract configurations (i.e., front/central/back articulation, open/half-open/half-close/close mouth) \cite{atal1982new} and the ``not a vowel" class.

\subsection{Vowel segmentation module}\label{sssec:Vowelseg}
We conduct vowel segmentation using a 250ms length, a choice that is informed by the mean vowel duration in American English (i.e., $\sim$200ms)\cite{jacewicz2007vowel} and has shown promising results in emotion recognition and formant frequency estimation tasks \cite{chen2021impact,dissen2019formant}. Each utterance is segmented into 250ms segments with a 125ms overlap, which allows to preserve the continuity of sequential information within the utterance.We use the FAVE aligner~\cite{yuan2008speaker} to find the vowel(s) included in each 250ms segment and obtain the timestamps of their boundaries. A given 250ms segment can be associated with one of three possible conditions: (1) no vowel or a single vowel fully included in the segment; (2) a single vowel partially included in the segment; or (3) more than one vowels included in the segment. The first case is very straightforward, and the corresponding segment will be labelled as /a/, /e/, /i/, /o/, /u/, or ``not a vowel". For the second case, a segment will be labeled as this vowel when more than 80\% of the vowel is included, or more than half of the segment is occupied by this vowel. Otherwise, it will be labeled as ``not a vowel''. For the third case, each vowel will be examined with the same requirements as in the second case. The label of this segment will be the first vowel in order of appearance in that segment that meets that condition, or ``not a vowel'' if none of the vowels qualifies. As a result, the outcome of this module is a set of 250ms audio segments and their corresponding vowel labels.

\begin{table*}[t]
\caption{The structure and hyper-parameters of the 2D convolutional neural network (CNN) used for vowel classification.}\vspace{-5pt}
\begin{tabular*}{\textwidth}{@{\extracolsep{\fill}}cccccc}
\hline
\begin{tabular}[c]{@{}c@{}}Layer \\ name\end{tabular} & Conv block 1 & Conv block 2 & Conv block 3 & Conv block 4 & Conv 5 \\ \hline
\begin{tabular}[c]{@{}c@{}}Parameter \\ Settings\end{tabular} & \begin{tabular}[c]{@{}c@{}}conv kernel (3, 1)\\ filter size 64\\ ReLU activation\\ pooling kernel (2, 1)\end{tabular} & \begin{tabular}[c]{@{}c@{}}conv kernel (3, 1)\\ filter size 64\\ ReLU activation\\ pooling kernel (2, 1)\end{tabular} & \begin{tabular}[c]{@{}c@{}}conv kernel (3, 1)\\ filter size 64\\ ReLU activation\\ pooling kernel (2, 1)\end{tabular} & \begin{tabular}[c]{@{}c@{}}conv kernel (3, 3)\\ filter size 64\\ ReLU activation\\ pooling kernel (2, 2)\end{tabular} & \begin{tabular}[c]{@{}c@{}}conv kernel \\(6, 6)\\ filter 6\end{tabular} \\
\begin{tabular}[c]{@{}c@{}}Output \\ shape\end{tabular} & (64, 63, 28) & (64, 30, 28) & (64, 14, 14) & (64, 6, 6) & (6, 1, 1) \\ \hline
\end{tabular*}
\label{tab: CNN}
\end{table*}

\subsection{Vowel classification module}\label{sssec:Vowelclass}
The goal of this module is to learn spectrotermporal energy variations of speech vowels. We first extract the log-Mel spectrogram for every 250ms segment that resulted from the vowel segmentation module. This feature is extracted using the Librosa Python library \cite{mcfee2015librosa} with the following parameters: 512-sample FFT window length, 128-sample hop length, and 128 Mel bands. The length of a 250ms window with a 16kHz sampling rate will be $4000$ samples. Without padding, the number of FFT window is $\lfloor(4000-512)/128\rfloor + 1=28$. As a result, each 250ms segments corresponds to a $(128, 28)$ spectrogram patch. The vowel classification module aims to learn a model $f$ with parameters $\boldsymbol{\theta}$ that takes as an input the log-Mel spectrogram $\mathbf{X}\in\mathbb{R}^{128\times28}$ and outputs the vowel class $\mathbf{y}\in\mathbb{R}^6$, such that $\mathbf{y}=f_{\boldsymbol{\theta}}(\boldsymbol{X})$.

The function $f$ is implemented via a 2D CNN, similar to prior work\cite{ma2016depaudionet,long2015fully}. Even though Ma {\it et al.} \cite{ma2016depaudionet} relied on a 1D CNN, which is commonly used in prior work \cite{srimadhur2020end}, we used the 2D CNN since it was more natural choice for the 2D input of speech spectrogram, thus contributing to the overall model explainability.  
The 2D CNN consists of four convolutional blocks, each including a convolutional, activation, batch normalization, and max pooling layer (Table \ref{tab: CNN}). Then a single convolutional layer is added to map the learned vowel-based embeddings into a vowel class. The model is implemented using Pytorch \cite{paszke2019pytorch} optimizing the cross-entropy loss with the following hyper-parameters: batch size 64, Adam optimizer with learning rate 0.001 and l2 regularization 0.001. A random over-sampling is applied to help with the balance the unbalanced distribution among vowel classes.

\begin{figure}[!tb]
  \centering
  \includegraphics[width=0.7\linewidth, trim = 3cm 0.5cm 10.5cm 3.5cm, clip=true, scale=0.5]{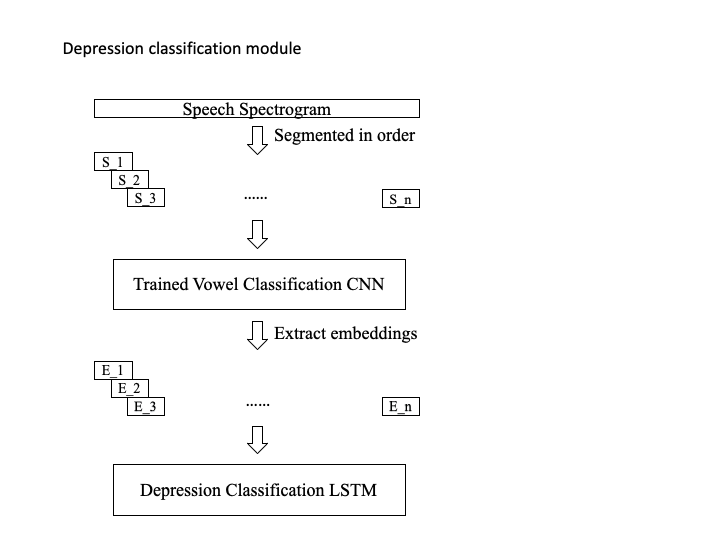}\vspace{-10pt}
 \caption{Visualization of the depression classification module) models.}\vspace{-5pt}
\label{fig:1}
\end{figure}

\subsection{Depression classification module}\label{sssec:Depressionclass}
This final module uses two types of vowel-based embeddings (Section \ref{sssec:Vowelclass}) as the input to a LSTM that models temporal dependencies within and across utterances. A comprehensive representation of this module is shown in Figure~\ref{fig:1}. The first type is the 6-dimensional output of the Conv 5 layer before the softmax operation (Table~\ref{tab: CNN}), referred to as ``Conv 5 vowel-based embedding". Since every channel of the Conv 5 output represents a vowel class, this embedding captures the activation of each vowel for a given 250ms segment. The second type of vowel-based embedding is the output of the Conv block 4 (Table~\ref{tab: CNN}), a 2304-dimensional vector after being flattened, referred to as ``Conv 4 vowel-based embedding". This is less interpretable compared to the Conv 5, but captures lower-level information closer to the speech spectrogram. Each LSTM model contains one LSTM layer with 128 hidden units, and one fully connected layer to estimate the likelihood of depression. The LSTM models are implemented using Pytorch \cite{paszke2019pytorch} and are optimized based on the cross-entropy loss using an Adam optimizer with 0.01 learning rate, 0.001 l2-norm regularization, and 32 batch size. Every depression sample is sampled twice to slightly balance the imbalanced distribution between the two classes without introducing speaker bias. The number of epochs is 7 and 39 for the LSTM trained based on the Conv 4 and Conv 5 vowel-based embeddings, respectively, due to their different input size. 

\subsection{Analysis of model explainability}\label{sssec:Explain}
We analyze two types of explainability: (1) a low-level explainability which corresponds to spectrotemporal patterns within each vowel that contribute most to the depression outcome; and (2) a high-level explainability that inspects the likelihood of the LSTM model in association to the depression outcome. The following discussion focuses on the system designed with the Conv 5 feature set, because its six dimensions represent the considered vowels, thus constributing to its interpretability.

In terms of low-level explainability, we examine the impact of the considered vowels on the decisions made by the proposed method. We evaluate the contribution of each vowel class to the final decision in the LSTM model. For this, we rely on a modified version of LIME \cite{ribeiro2016should}, a method widely used for explaining the decision-making process of ML systems. LIME randomly perturbs part of the original input and observes the change in the model's output, while also tracking similarities between the original and perturbed samples. By fitting a ridge regression model on these perturbed samples and the original model's output, the weight of the model indicates the significance of each part of the input. We customize LIME to account for time-series using an existing implementation \cite{metzenthin2021lime} (Algorithm \ref{Alg: LIME}). We restrict the perturbation to happen on the channel level, which corresponds to one of the six dimensions of the Conv 5 vowel-based embedding. Channel perturbation is conducted by replacing randomly selected values of a channel with values generated from a uniform distribution, whose range is the same as the range of the original non-perturbed data in that channel. The impact of a channel on a sample's classification decision is measured by the absolute value of weight $\mathbf{w}$ returned by the algorithm. The channels that correspond to the largest absolute values $\mathbf{w}$ will have the highest impact to the decision-making process.

In terms of high-level explainability, we observe the temporal change of decisions provided by our method. More specifically, we feed the hidden state in every time-step into the fully connected layer of the LSTM and plot the decision change over time. Since we are using a short time step of 250ms length, the corresponding decisions are noisy, thus we smooth these decisions over a 4s window. We next observe the consistency of the plot in association to the final depression decision. Plotting the decision change is inherently a similar idea as the one proposed by Strobelt {\it et al.} in LSTMVis, but more straightforward and easier to observe, since Strobelt {\it et al.} relied on the hidden states of the LSTM, whose dimensionality is too high (i.e., 128) in our case.

\begin{algorithm}[!hbt]
\caption{Modified version of LIME}
\begin{algorithmic}
\State \textbf{Input:} 
\State --- LSTM classifier $f$, input sample $\mathbf{X}$, similarity kernel $s$, number of perturbed samples $N$
\State \textbf{Output:} weight $\mathbf{w}$ of the fitted linear model
\State $Z \gets \{\}$ \Comment{Initialize the perturbed set}
\For{i from 0 to $N$} \Comment{generate $N$ perturbed samples}
    \State $\mathbf{X'} = \mathbf{X}$
    \State $perturb\_matrix = [1, 1, 1, 1, 1, 1]$ \Comment{perturbation per channel}
    \State Randomly decide the number of channels ($m$) to perturb
    \For{j from 0 to $m$}
        \State Randomly select an unperturbed channel index $n$
        \State Generate $r$ via a uniform distribution
        \State Replace the channel $n$ of $\mathbf{X'}$ with $r$
        \State $perturb\_matrix[n]=0$
    \EndFor
    \State $Z \gets Z \cup \{perturb\_matrix, f(\mathbf{X'}), s(\mathbf{X}, \mathbf{X'})\}$
\EndFor
\State $Z \gets Z \cup \{[1, 1, 1, 1, 1, 1], f(\mathbf{X}), 0\}$ \Comment{add original sample}

\State $\mathbf{w} \gets RidgeRegress(Z)$ \Comment{$perturb\_matrix$ as feature, $f(\mathbf{X'})$ as target, $s(\mathbf{X}, \mathbf{X'})$ as sample weight}
\State \Return{$\mathbf{w}$} 
\end{algorithmic}
\label{Alg: LIME}
\end{algorithm}

\section{Experiments}

\subsection{Data description}
We used the Wizard-of-Oz part of the Distress Analysis Interview Corpus (DAIC-WoZ) \cite{gratch2014distress}. This contains 142 clinical interviews between a participant and a virtual agent operated in a Wizard-of-Oz paradigm, used to elicit consistent responses across participants. Each participant also filled out the Patient Health Questionnaire (PHQ-8) \cite{gilbody2007screening}, which serves as an indicator of MH condition. A PHQ-8 score larger or equal to 10 is considered a threshold for depression, while a PHQ-8 score less than 10 is considered to come from a healthy participant. Among these 142 participants, 107 participants are included in the training set (30 depressed and 77 non-depressed), and 35 are in the development set (12 depressed and 23 non-depressed), similar to the original AVEC 2016 challenge \cite{valstar2016avec}. The performance is reported on the development set to ensure a compatible baseline from previous work and since the labels of the test set of the data are unavailable.

\subsection{Baseline methods}
The first baseline is DepAudioNet \cite{ma2016depaudionet}, which is the closest to our method. The difference between the two is that DepAudioNet does not use vowel information, therefore the CNN and LSTM are trained in an end-to-end manner for depression classification. The second baseline is the SpeechFormer \cite{chen2022speechformer}, a more complex architecture that uses a hierarchical framework for modeling spectral variations within a speech frame, phoneme, word, and utterance, as well as across these. The third baseline comes from the original model provided in the AVEC 2016 challenge \cite{valstar2016avec} and contains a SVM trained on a 74-dimensional acoustic features at the frame-level. Decisions at the frame-level are combined via majority voting.

\begin{table*}[t]
\caption{Precision/Recall/F1-score of vowel classification with different confidence thresholds (CT) and the corresponding percentage of segments included for calculating the performance metrics. The number of samples for each vowel class is in parentheses.}\vspace{-5pt}
\scriptsize
\begin{tabular*}{\textwidth}{@{\extracolsep{\fill}}ccccccccc}
\hline
CT & \% segments & /a/ ($N=$ 37013) & /e/ ($N=$ 10905) & /i/ ($N=$ 13582) & /o/ ($N=$ 4180) & /u/ ($N=$ 2969) & Not vowel ($N=$ 54632) & macro\\ \hline
0 & 100 & 0.725/0.571/0.639 & 0.306/0.544/0.392 & 0.509/0.5/0.505 & 0.294/0.487/0.367 & 0.261/0.508/0.345 & 0.748/0.667/0.705 & 0.474/0.546/0.492\\
0.3 & 99 & 0.728/0.574/0.642 & 0.309/0.547/0.395 & 0.513/0.504/0.508 & 0.297/0.492/0.37 & 0.265/0.512/0.349 & 0.75/0.67/0.708 & 0.477/0.55/0.495 \\
0.6 & 60 & 0.82/0.698/0.754 & 0.421/0.652/0.512 & 0.642/0.619/0.63 & 0.371/0.621/0.465 & 0.361/0.685/0.473 & 0.848/0.778/0.811 & 0.577/0.676/0.607 \\
0.9 & 19 & 0.926/0.907/0.916 & 0.63/0.783/0.698 & 0.812/0.783/0.797 & 0.539/0.803/0.645 & 0.54/0.935/0.685 & 0.972/0.887/0.928 & 0.737/0.85/0.778 \\ \hline
\end{tabular*} \label{tab: vowel result}
\end{table*}

\begin{table}[t]\scriptsize
\caption{Depression classification performance for the proposed and baseline methods.}\vspace{-5pt}
\begin{center}
\begin{tabular}{ccccc}
\hline
Method & Precision & Recall & F1 & Macro F1 \\ \hline
AVEC 2016 baseline \cite{valstar2016avec} & 0.32 & 0.85 & 0.46 & 0.57 \\
DepAudioNet (2016) \cite{ma2016depaudionet} & 0.35 & 1.0 & 0.52 & 0.61 \\
SpeechFormer (2022) \cite{chen2022speechformer} & - & - & - & 0.69 \\
Conv 5 vowel-based embedding & 0.5 & 0.75 & \textbf{0.6} & 0.65 \\
Conv 4 vowel-based embedding & \textbf{0.83} & 0.42 & 0.56 & \textbf{0.70} \\ \hline
\end{tabular}
\end{center}
\label{tab: depression result}
\end{table}

\begin{figure*}[!htb]
  \centering
  \begin{minipage}{0.26\linewidth}
  \includegraphics[trim = 0.7cm 0.2cm 1.6cm 1.3cm, clip=true, scale=0.36]{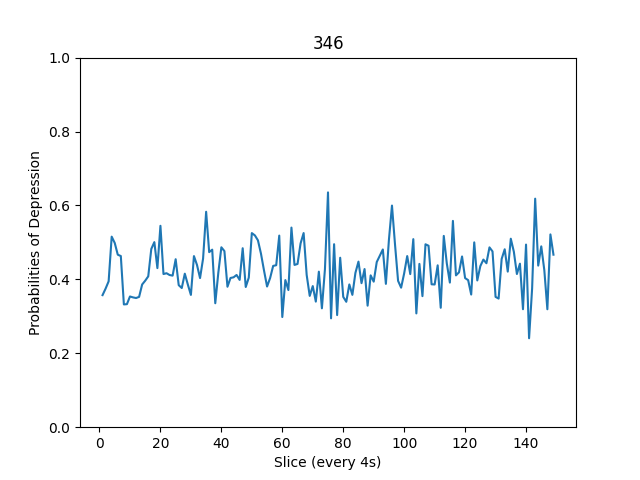}
  \centerline{\tiny {Participant ID = 346; Prediction=1; Label=1}}
  \end{minipage}
  \hfill
  \begin{minipage}{0.24\linewidth}
  \includegraphics[trim = 1.8cm 0.2cm 1.6cm 1.3cm, clip=true, scale=0.36]{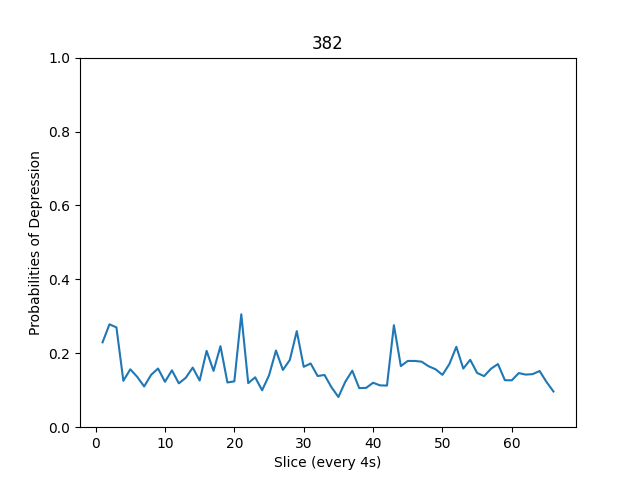}
  \centerline{\tiny {Participant ID = 382; Prediction=0; Label=0}}
  \end{minipage}
  \hfill
  \begin{minipage}{0.24\linewidth}
  \includegraphics[trim = 1.8cm 0.2cm 1.6cm 1.3cm, clip=true, scale=0.36]{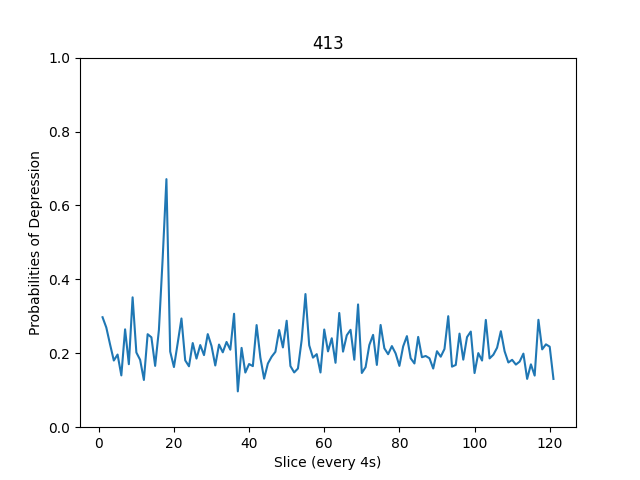}
  \centerline{\tiny {Participant ID = 413; Prediction=0; Label=1}}
  \end{minipage}
  \hfill
  \begin{minipage}{0.24\linewidth}
  \includegraphics[trim = 1.8cm 0.2cm 1.6cm 1.3cm, clip=true, scale=0.36]{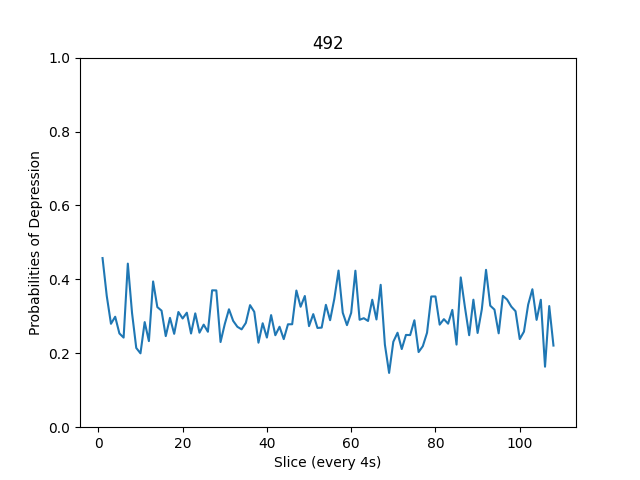}
  \centerline{\tiny {Participant ID = 346; Prediction=1; Label=0}}
  \end{minipage}
\caption{Change in depression probability over time for various participants with different depression labels and predictions.}
\label{fig: pred}
\end{figure*}

\subsection{Results} \label{sec:results}

We first report the performance of the vowel classification model (Section~\ref{sssec:Vowelclass}), which will contribute to understanding the ability of the CNN to reliably capture vowel-based descriptors. Vowel classification is evaluated via precision, recall, and F1-score for each vowel class, and the corresponding macro-averages (i.e., average metric per class). Even though we have adopted a comprehensive labeling protocol for labeling every 250ms segment, a segment can be inherently noisy, since it can contain more than one vowels or contain both a vowel and a non-vowel. As a result, we report the model performance over different confidence thresholds (CT), where confidence is defined as the highest value of the softmax output of the CNN (Table~\ref{tab: CNN}). For each CT, the samples with a confidence value lower than the predefined CT are not included when calculating the phoneme classification measures. Setting a higher CT results in excluding noisy samples from the evaluation. The aforementioned performance scores are calculated across the 250ms segments extracted from the development set. Results are shown with four CTs (i.e., 0, 0.3, 0.6, 0.9), for which we also report the percentage of considered 250ms segments that are included in the evaluation (Table \ref{tab: vowel result}). As expected, vowel classification accuracies improve when lowering the CT. Some vowels, such as /a/ and /i/, are easier to classify compared to others. The ``not a vowel" class also appears to perform well. Our model depicts comparable performance in common vowels (e.g., /a/) as in previous work \cite{sabzi2019comparative, wesker2005oldenburg}. However, previous work often uses acted speech, such as the Oldenburg Logatome corpus \cite{sabzi2019comparative, wesker2005oldenburg}, that is rich in both consonants and vowels. Factors such as an increased number of speakers, spontaneous speech, and imbalanced vowel distribution add more challenges to our task.

Next we present the depression classification results (Table \ref{tab: depression result}). Evaluation metrics include the precision, recall, and F1-score for the depression class, as well as the macro F1-score, that is the average of F1-scores separately computed for the depression and no-depression classes, which is the standard evaluation method for the AVEC 2016 challenge \cite{valstar2016avec}. The proposed method that uses the Conv 5 vowel-based feature embedding outperforms both the AVEC 2016 baseline and the DepAudioNet in terms of depression F1-score. The Conv 4 vowel-based feature embedding further leads to the highest precision and macro F1-score, as compared to all three baseline systems. Although the SpeechFormer appears to be very close to the proposed approach, the SpeechFormer takes advantage of additional information beyond the phoneme-level (i.e., word and utterance level), which renders it a challenging baseline. However, our results, and especially the comparison between the proposed approach and the DepAudioNet, demonstrate the benefit of modeling vowel-based information for the task of depression classification.

We analyze the weights $\mathbf{w}$ learned from LIME to understand the low-level explainability of the proposed model (Section~\ref{sssec:Explain}). We compute the absolute values of weights $\mathbf{w}$ estimated by LIME for each vowel over the utterances of the 35 participants in the development set. For each participant, we calculate the number of times the absolute value of the weight of each vowel ranked first, second, third, etc. We sum this occurrence and present the result in Table \ref{tab: lime result}. We observe that the class ``Not a vowel'' is mostly at the 5th or 6th ranking position, which indicates that spectrotemporal information corresponding to non-vowel segments is not as important as the vowel-based information~\cite{shimizu2005chaos, scherer2015self, vlasenko2017implementing, stasak2019investigation}. We also observe that embeddings learned for vowel /a/ tend to depict the highest importance weights, which might be also due to the fact that /a/ is the most common vowel in the data.

We finally plot the temporal change in decisions provided by the LSTM model for various participants (Figure \ref{fig: pred}). We observe that for participant 346 with depression label, the corresponding likelihood of the model is closer to 0.5. Participant 382 does not have depression and the corresponding likelihood is around to 0.2. In addition, the probability change of depression for each segment is stable for both participants without any bursts, which indicates that the model can estimate the correct output without any significant fluctuations over time. On the other hand, we observe some burst in the corresponding time-series of decisions for participant 413, who is depressed, but the model estimates no-depression. While observing the prediction change for this participant, we find an abrupt probability increase at around slice 20 (80 seconds). This participant is an ambiguous example, since their corresponding PHQ-8 score is 10, which is exactly at the boundary between depression and non-depression. In the last example, we show the depression probability change of participant 492, who does not have depression but was classified as depressed. As the majority of probabilities for this participant is below 0.5, it would be reasonable for a user to consider that this prediction is a false positive.

\begin{table}[t]
\caption{The number of samples on different impact level for the development set.}
\begin{tabular*}{\linewidth}{@{\extracolsep{\fill}}ccccccc}
\hline
Vowel & 1st & 2nd & 3rd & 4th & 5th & 6th \\ \hline
/a/ & 9 & 11 & 2 & 1 & 6 & 6 \\
/e/ & 2 & 5 & 7 & 14 & 3 & 4 \\
/i/ & 7 & 7 & 6 & 5 & 7 & 3 \\
/o/ & 2 & 5 & 6 & 8 & 8 & 5 \\
/u/ & 12 & 5 & 8 & 2 & 2 & 6 \\
Not a vowel & 3 & 2 & 6 & 5 & 8 & 11 \\ \hline
\end{tabular*}
\label{tab: lime result}
\end{table}

\section{Discussion} \label{Sec: dis}
This paper presented a novel approach to detect depression from speech by modeling spectrotemporal information around the speech vowels. The proposed approach integrated low-level vowel-based information on a high-level representation to take the final MH decision. The explainability of the model was further observed at the low-level of the vowel formation and the high-level of the temporal sequence of decisions for consecutive speech segments. Such knowledge-driven approaches have the potential to yield explainable AI solutions that could potentially support clinicians and MH practitioners in their decision-making by pointing out to specific variations in the speech signal that might be relevant to MH degradation (i.e., otherwise not easily perceivable by humans), and assisting in better understanding high-level information regarding the temporal evolution of phonetic variation in speech. As part of our future work, we plan to integrate additional contextual and semantic information in the model by explicitly modeling words and the utterances within the data, as well as integrating linguistic information with the acoustic features. We further plan to conduct perceptual experiments in which the explainability, usability, and trustworthiness of the proposed method will be evaluated by MH clinicians. Finally, we will address the imbalanced nature of DAI-WOZ dataset by appropriate data augmentation methods (e.g., random sampling of positive samples, oversampling of vowel-based information).

\section{Conclusions}
In this paper, we introduced a new knowledge-based deep learning method for detecting depression. A CNN model is trained to model the 250ms vowel-based segments for the task of vowel classification. We used an LSTM model to model the temporal evolution of learned vowel-based embeddings and conduct the final depression classification task. We evaluated our method on the DAIC-WOZ dataset and our model outperforms various baseline methods. Explainability analysis of the low-level information captured by the proposed model suggests that vowels play an important role for the depression outcome. Explainability of the high-level information capturing the segment-by-segment decisions of the model was further analyzed in terms of various examples of participants. As part of our future work, we plan to apply more advanced data augmentation method for both the CNN and LSTM models and work toward improving the explainability of the LSTM model by taking into account its hidden layers.

\bibliographystyle{IEEEtran}
\bibliography{refs}

\begin{thebibliography}{10}
\providecommand{\url}[1]{#1}
\csname url@samestyle\endcsname
\providecommand{\newblock}{\relax}
\providecommand{\bibinfo}[2]{#2}
\providecommand{\BIBentrySTDinterwordspacing}{\spaceskip=0pt\relax}
\providecommand{\BIBentryALTinterwordstretchfactor}{4}
\providecommand{\BIBentryALTinterwordspacing}{\spaceskip=\fontdimen2\font plus
\BIBentryALTinterwordstretchfactor\fontdimen3\font minus
  \fontdimen4\font\relax}
\providecommand{\BIBforeignlanguage}[2]{{%
\expandafter\ifx\csname l@#1\endcsname\relax
\typeout{** WARNING: IEEEtran.bst: No hyphenation pattern has been}%
\typeout{** loaded for the language `#1'. Using the pattern for}%
\typeout{** the default language instead.}%
\else
\language=\csname l@#1\endcsname
\fi
#2}}
\providecommand{\BIBdecl}{\relax}
\BIBdecl

\bibitem{who_depression}
\BIBentryALTinterwordspacing
``World health organization depression information page.'' [Online]. Available:
  \url{https://www.who.int/news-room/fact-sheets/detail/depression}
\BIBentrySTDinterwordspacing

\bibitem{katon2013depression}
W.~J. Katon, ``Depression research in under-resourced populations: An
  academic--community partnership,'' \emph{Journal of general internal
  medicine}, vol.~28, no.~10, pp. 1255--1257, 2013.

\bibitem{mongelli2020challenges}
F.~Mongelli, P.~Georgakopoulos, and M.~T. Pato, ``Challenges and opportunities
  to meet the mental health needs of underserved and disenfranchised
  populations in the united states,'' \emph{Focus}, vol.~18, no.~1, pp. 16--24,
  2020.

\bibitem{mundt2007voice}
J.~C. Mundt, P.~J. Snyder, M.~S. Cannizzaro, K.~Chappie, and D.~S. Geralts,
  ``Voice acoustic measures of depression severity and treatment response
  collected via interactive voice response (ivr) technology,'' \emph{Journal of
  neurolinguistics}, vol.~20, no.~1, pp. 50--64, 2007.

\bibitem{spruijt2014dynamic}
D.~Spruijt-Metz and W.~Nilsen, ``Dynamic models of behavior for just-in-time
  adaptive interventions,'' \emph{IEEE Pervasive Computing}, vol.~13, no.~3,
  pp. 13--17, 2014.

\bibitem{james2022preparing}
C.~A. James, R.~M. Wachter, and J.~O. Woolliscroft, ``Preparing clinicians for
  a clinical world influenced by artificial intelligence,'' \emph{JAMA}, vol.
  327, no.~14, pp. 1333--1334, 2022.

\bibitem{ginestra2019clinician}
J.~C. Ginestra, H.~M. Giannini, W.~D. Schweickert, L.~Meadows, M.~J. Lynch,
  K.~Pavan, C.~J. Chivers, M.~Draugelis, P.~J. Donnelly, B.~D. Fuchs
  \emph{et~al.}, ``Clinician perception of a machine learning-based early
  warning system designed to predict severe sepsis and septic shock,''
  \emph{Critical care medicine}, vol.~47, no.~11, p. 1477, 2019.

\bibitem{atal1982new}
B.~Atal and J.~Remde, ``A new model of lpc excitation for producing
  natural-sounding speech at low bit rates,'' in \emph{ICASSP'82. IEEE
  International Conference on Acoustics, Speech, and Signal Processing},
  vol.~7.\hskip 1em plus 0.5em minus 0.4em\relax IEEE, 1982, pp. 614--617.

\bibitem{murphy1999emotional}
F.~Murphy, B.~Sahakian, J.~Rubinsztein, A.~Michael, R.~Rogers, T.~Robbins, and
  E.~Paykel, ``Emotional bias and inhibitory control processes in mania and
  depression,'' \emph{Psychological medicine}, vol.~29, no.~6, pp. 1307--1321,
  1999.

\bibitem{shimizu2005chaos}
T.~Shimizu, N.~Furuse, T.~Yamazaki, Y.~Ueta, T.~Sato, and S.~Nagata, ``Chaos of
  vowel/a/in japanese patients with depression: a preliminary study,''
  \emph{Journal of occupational health}, vol.~47, no.~3, pp. 267--269, 2005.

\bibitem{scherer2015self}
S.~Scherer, G.~M. Lucas, J.~Gratch, A.~S. Rizzo, and L.-P. Morency,
  ``Self-reported symptoms of depression and {PTSD} are associated with reduced
  vowel space in screening interviews,'' \emph{IEEE Transactions on Affective
  Computing}, vol.~7, no.~1, pp. 59--73, 2015.

\bibitem{vlasenko2017implementing}
B.~Vlasenko, H.~Sagha, N.~Cummins, and B.~Schuller, ``Implementing
  gender-dependent vowel-level analysis for boosting speech-based depression
  recognition,'' 2017.

\bibitem{stasak2019investigation}
B.~Stasak, J.~Epps, and R.~Goecke, ``An investigation of linguistic stress and
  articulatory vowel characteristics for automatic depression classification,''
  \emph{Computer Speech \& Language}, vol.~53, pp. 140--155, 2019.

\bibitem{quatieri2012vocal}
T.~F. Quatieri and N.~Malyska, ``Vocal-source biomarkers for depression: A link
  to psychomotor activity,'' in \emph{Thirteenth annual conference of the
  international speech communication association}, 2012.

\bibitem{williamson2013vocal}
J.~R. Williamson, T.~F. Quatieri, B.~S. Helfer, R.~Horwitz, B.~Yu, and D.~D.
  Mehta, ``Vocal biomarkers of depression based on motor incoordination,'' in
  \emph{Proceedings of the 3rd ACM international workshop on Audio/visual
  emotion challenge}, 2013, pp. 41--48.

\bibitem{cummins2015analysis}
N.~Cummins, V.~Sethu, J.~Epps, S.~Schnieder, and J.~Krajewski, ``Analysis of
  acoustic space variability in speech affected by depression,'' \emph{Speech
  Communication}, vol.~75, pp. 27--49, 2015.

\bibitem{valstar2013avec}
M.~Valstar, B.~Schuller, K.~Smith, F.~Eyben, B.~Jiang, S.~Bilakhia,
  S.~Schnieder, R.~Cowie, and M.~Pantic, ``Avec 2013: the continuous
  audio/visual emotion and depression recognition challenge,'' in
  \emph{Proceedings of the 3rd ACM international workshop on Audio/visual
  emotion challenge}, 2013, pp. 3--10.

\bibitem{cohn2018multimodal}
J.~F. Cohn, N.~Cummins, J.~Epps, R.~Goecke, J.~Joshi, and S.~Scherer,
  ``Multimodal assessment of depression from behavioral signals,'' \emph{The
  Handbook of Multimodal-Multisensor Interfaces: Signal Processing,
  Architectures, and Detection of Emotion and Cognition-Volume 2}, pp.
  375--417, 2018.

\bibitem{ma2016depaudionet}
X.~Ma, H.~Yang, Q.~Chen, D.~Huang, and Y.~Wang, ``Depaudionet: An efficient
  deep model for audio based depression classification,'' in \emph{Proceedings
  of the 6th international workshop on audio/visual emotion challenge}, 2016,
  pp. 35--42.

\bibitem{muzammel2020audvowelconsnet}
M.~Muzammel, H.~Salam, Y.~Hoffmann, M.~Chetouani, and A.~Othmani,
  ``Audvowelconsnet: A phoneme-level based deep cnn architecture for clinical
  depression diagnosis,'' \emph{Machine Learning with Applications}, vol.~2, p.
  100005, 2020.

\bibitem{saidi2020hybrid}
A.~Saidi, S.~B. Othman, and S.~B. Saoud, ``Hybrid cnn-svm classifier for
  efficient depression detection system,'' in \emph{2020 4th International
  Conference on Advanced Systems and Emergent Technologies (IC\_ASET)}.\hskip
  1em plus 0.5em minus 0.4em\relax IEEE, 2020, pp. 229--234.

\bibitem{srimadhur2020end}
N.~Srimadhur and S.~Lalitha, ``An end-to-end model for detection and assessment
  of depression levels using speech,'' \emph{Procedia Computer Science}, vol.
  171, pp. 12--21, 2020.

\bibitem{zhao2020hierarchical}
Z.~Zhao, Z.~Bao, Z.~Zhang, N.~Cummins, H.~Wang, and B.~Schuller, ``Hierarchical
  attention transfer networks for depression assessment from speech,'' in
  \emph{ICASSP 2020-2020 IEEE International Conference on Acoustics, Speech and
  Signal Processing (ICASSP)}.\hskip 1em plus 0.5em minus 0.4em\relax IEEE,
  2020, pp. 7159--7163.

\bibitem{scherer2015reduced}
S.~Scherer, L.-P. Morency, J.~Gratch, and J.~Pestian, ``Reduced vowel space is
  a robust indicator of psychological distress: A cross-corpus analysis,'' in
  \emph{2015 IEEE International Conference on Acoustics, Speech and Signal
  Processing (ICASSP)}.\hskip 1em plus 0.5em minus 0.4em\relax IEEE, 2015, pp.
  4789--4793.

\bibitem{jacewicz2007vowel}
E.~Jacewicz, R.~A. Fox, and J.~Salmons, ``Vowel duration in three american
  english dialects,'' \emph{American Speech}, vol.~82, no.~4, pp. 367--385,
  2007.

\bibitem{chen2021impact}
S.~Chen, M.~Zhang, X.~Yang, Z.~Zhao, T.~Zou, and X.~Sun, ``The impact of
  attention mechanisms on speech emotion recognition,'' \emph{Sensors},
  vol.~21, no.~22, p. 7530, 2021.

\bibitem{dissen2019formant}
Y.~Dissen, J.~Goldberger, and J.~Keshet, ``Formant estimation and tracking: A
  deep learning approach,'' \emph{The Journal of the Acoustical Society of
  America}, vol. 145, no.~2, pp. 642--653, 2019.

\bibitem{yuan2008speaker}
J.~Yuan, M.~Liberman \emph{et~al.}, ``Speaker identification on the scotus
  corpus,'' \emph{Journal of the Acoustical Society of America}, vol. 123,
  no.~5, p. 3878, 2008.

\bibitem{mcfee2015librosa}
B.~McFee, C.~Raffel, D.~Liang, D.~P. Ellis, M.~McVicar, E.~Battenberg, and
  O.~Nieto, ``librosa: Audio and music signal analysis in python,'' in
  \emph{Proceedings of the 14th python in science conference}, vol.~8.\hskip
  1em plus 0.5em minus 0.4em\relax Citeseer, 2015, pp. 18--25.

\bibitem{long2015fully}
J.~Long, E.~Shelhamer, and T.~Darrell, ``Fully convolutional networks for
  semantic segmentation,'' in \emph{Proceedings of the IEEE conference on
  computer vision and pattern recognition}, 2015, pp. 3431--3440.

\bibitem{paszke2019pytorch}
A.~Paszke, S.~Gross, F.~Massa, A.~Lerer, J.~Bradbury, G.~Chanan, T.~Killeen,
  Z.~Lin, N.~Gimelshein, L.~Antiga \emph{et~al.}, ``Pytorch: An imperative
  style, high-performance deep learning library,'' \emph{Advances in neural
  information processing systems}, vol.~32, 2019.

\bibitem{ribeiro2016should}
M.~T. Ribeiro, S.~Singh, and C.~Guestrin, ``" why should i trust you?"
  explaining the predictions of any classifier,'' in \emph{Proceedings of the
  22nd ACM SIGKDD international conference on knowledge discovery and data
  mining}, 2016, pp. 1135--1144.

\bibitem{metzenthin2021lime}
E.~Metzenthin, ``Lime for time,''
  \url{https://github.com/emanuel-metzenthin/Lime-For-Time}, 2021.

\bibitem{gratch2014distress}
J.~Gratch, R.~Artstein, G.~Lucas, G.~Stratou, S.~Scherer, A.~Nazarian, R.~Wood,
  J.~Boberg, D.~DeVault, S.~Marsella \emph{et~al.}, ``The distress analysis
  interview corpus of human and computer interviews,'' in \emph{Proceedings of
  the Ninth International Conference on Language Resources and Evaluation
  (LREC'14)}, 2014, pp. 3123--3128.

\bibitem{gilbody2007screening}
S.~Gilbody, D.~Richards, S.~Brealey, and C.~Hewitt, ``Screening for depression
  in medical settings with the patient health questionnaire (phq): a diagnostic
  meta-analysis,'' \emph{Journal of general internal medicine}, vol.~22,
  no.~11, pp. 1596--1602, 2007.

\bibitem{valstar2016avec}
M.~Valstar, J.~Gratch, B.~Schuller, F.~Ringeval, D.~Lalanne, M.~Torres~Torres,
  S.~Scherer, G.~Stratou, R.~Cowie, and M.~Pantic, ``Avec 2016: Depression,
  mood, and emotion recognition workshop and challenge,'' in \emph{Proceedings
  of the 6th international workshop on audio/visual emotion challenge}, 2016,
  pp. 3--10.

\bibitem{chen2022speechformer}
W.~Chen, X.~Xing, X.~Xu, J.~Pang, and L.~Du, ``Speechformer: A hierarchical
  efficient framework incorporating the characteristics of speech,''
  \emph{arXiv preprint arXiv:2203.03812}, 2022.

\bibitem{sabzi2019comparative}
A.~Sabzi~Shahrebabaki, A.~S. Imran, N.~Olfati, and T.~Svendsen, ``A comparative
  study of deep learning techniques on frame-level speech data
  classification,'' \emph{Circuits, Systems, and Signal Processing}, vol.~38,
  no.~8, pp. 3501--3520, 2019.

\bibitem{wesker2005oldenburg}
T.~Wesker, B.~Meyer, K.~Wagener, J.~Anem{\"u}ller, A.~Mertins, and
  B.~Kollmeier, ``Oldenburg logatome speech corpus (ollo) for speech
  recognition experiments with humans and machines,'' in \emph{Ninth European
  Conference on Speech Communication and Technology}.\hskip 1em plus 0.5em
  minus 0.4em\relax Citeseer, 2005.

\end{thebibliography}

\end{document}